# Comparative Evaluation of Explainable Machine Learning Versus Linear Regression for Predicting County-Level Lung Cancer Mortality Rate in the United States

Soheil Hashtarkhani, PhD[1] ; Brianna M. White, MPH[1] ; Benyamin Hoseini, PhD[2]; David L. Schwartz, MD[3] ; and Arash Shaban-Nejad, PhD, MPH[1]

## ABSTRACT

**PURPOSE** Lung cancer (LC) is a leading cause of cancer-related mortality in the United States. Accurate prediction of LC mortality rates is crucial for guiding targeted interventions and addressing health disparities. Although traditional regression-based models have been commonly used, explainable machine learning models may offer enhanced predictive accuracy and deeper insights into the factors influencing LC mortality.

**METHODS** This study applied three models—random forest (RF), gradient boosting regression (GBR), and linear regression (LR)—to predict county-level LC mortality rates across the United States. Model performance was evaluated using R-squared and root mean squared error (RMSE). Shapley Additive Explanations (SHAP) values were used to determine variable importance and their directional impact. Geographic disparities in LC mortality were analyzed through Getis-Ord (Gi*) hotspot analysis.

**RESULTS** The RF model outperformed both GBR and LR, achieving an $R^2$ value of 41.9% and an RMSE of 12.8. SHAP analysis identified smoking rate as the most important predictor, followed by median home value and the percentage of the Hispanic ethnic population. Spatial analysis revealed significant clusters of elevated LC mortality in the mid-eastern counties of the United States.

**CONCLUSION** The RF model demonstrated superior predictive performance for LC mortality rates, emphasizing the critical roles of smoking prevalence, housing values, and the percentage of Hispanic ethnic population. These findings offer valuable actionable insights for designing targeted interventions, promoting screening, and addressing health disparities in regions most affected by LC in the United States.

## INTRODUCTION

Cancer, with its multifaceted impact on public health and society, remains one of the most significant challenges in contemporary medicine. Among its various forms, lung cancer (LC) stands out as a major contributor to cancer-related morbidity and mortality worldwide.[1] In the United States, LC has been a leading cause of cancer-related mortality in both men and women over several decades.[2] With a mortality rate exceeding that of breast, prostate, and pancreatic cancers combined, and more than 2.5 times higher than colorectal cancer (the second leading cause of cancer mortality in the country), LC demands urgent attention.[3] Although advances in technologies and screening have led to a decline in new LC diagnoses since the early 2010s, the disease remains a critical threat to public health, with a 5-year relative survival rate of just 26.6%, as reported by the American Lung Association.[4] Given these statistics, there is a clear and pressing need for effective predictive models to guide interventions and optimize resource allocation.

Numerous studies have used statistical and geospatial techniques to model the association between risk factors and county-level LC mortality.[5-8] At the population level, a variety of socioeconomic, environmental, and demographic determinants influence the incidence and mortality rates of LC.[9-11] For instance, Li et al[5] applied simultaneous autoregressive models, which account for spatial interdependencies, to assess the associations between the Environmental Quality Index (EQI) and US county-level LC mortality. The EQI incorporates five domain-specific indices (air, water, land, built environment, and sociodemographic factors) into a single composite index. Their findings highlighted a

## CONTEXT

**Key Objective**

Do ensemble machine learning (ML) models outperform traditional linear regression (LR) in predicting county-level lung cancer (LC) mortality rates, and what geographic and socioeconomic factors drive these predictions?

**Knowledge Generated**

Both random forest and gradient boosting regression outperformed LR in capturing nonlinear relationships. Shapley Additive Explanations analysis confirmed smoking as the strongest LC predictor while suggesting protective effects of higher median home values and Hispanic population percentage.

**Relevance (P.-M. Putora)**

This study investigated LC mortality disparities in the United States, using and comparing different ML methodologies.*

*Relevance section written by *JCO Clinical Cancer Informatics* Associate Editor Paul-Martin Putora, MD, PhD, MA.

significant correlation between lower EQI and increased LC mortality, with a stronger effect size observed in females compared with males. Similarly, an investigation within the United States[6] found an increased risk of county-level LC mortality associated with higher levels of environmental carcinogen releases, demonstrated through linear regression (LR). Notably, although these associations were observed across gender and racial groups (including Whites and African Americans), the strength of these relationships was more evident among African American cohorts. These studies underscore the importance of considering gender and environmental factors when developing strategies to reduce LC mortality. Moreover, a notable geospatial variation in LC mortality rates is evident, both interstate and intrastate, extending down to the county level.[7,12] However, it is important to note that these studies primarily rely on conventional linear models, which may not adequately capture the complex interactions and nonlinear dynamics between LC mortality and its risk factors.[13,14]

Although traditional regression-based models have historically played a fundamental role in cancer outcome prognostication,[15,16] the emergence of explainable machine learning (ML) techniques has introduced new opportunities to improve predictive accuracy and enhance our understanding of complex disease dynamics.[17,18] These advanced techniques have revealed higher-order nonlinear interactions among variables, leading to more robust and reliable predictions.[13] For example, Lee et al[19] conducted a comprehensive analysis using both regression and ML techniques, to examine the impacts of background radiation levels, PM2.5 exposure, and various sociobehavioral factors on county-level LC incidence rates in the United States. Their findings showed the superiority of the random forest (RF) algorithm over traditional regression methods. Although cancer research has made significant progress in assessing the impacts of social determinants of health on disease risk,[20,21] ML techniques can further advance our understanding. Given the increasing availability of data and advances in ML techniques, there is a significant opportunity to more accurately predict and identify the factors influencing LC mortality. These insights can inform targeted treatment strategies and interventions tailored to specific geographic areas.

Ensemble learning (EL), a powerful ML technique, enhances predictive accuracy by leveraging the collective intelligence of multiple models.[13,22] Although some studies suggest that EL models may not outperform logistic regression models in predicting mortality in emergency departments,[18,23] EL proves to be a compelling alternative to traditional LR for forecasting county-level LC mortality rates, for several reasons: (1) The complex interplay of socioeconomic, environmental, and health care factors influencing cancer outcomes requires a nuanced understanding that EL's capacity to capture complex relationships can provide. (2) County-level health data sets often suffer from sparse, heterogeneous, or inconsistent data, posing challenges for conventional models. EL's resilience to such data irregularities enables robust predictions despite these obstacles. (3) By combining diverse models that perform well on different subsets of the data, EL algorithms improve prediction accuracy, which is essential for addressing the heterogeneity observed in cancer epidemiology. (4) The adaptability of EL to accommodate varied hyperparameter configurations is crucial for customizing models to the unique characteristics of different counties and their populations.

Although the potential benefits of EL techniques in predictive modeling are widely acknowledged,[13,24] their evaluation—particularly in comparison with traditional LR models—remains limited, especially in the context of LC. Therefore, the primary objective of this study is to compare the predictive performance of EL models with LR models for county-level LC mortality in the United States. Additionally, the study aims to explore geographical disparities in LC mortality rates and associated community-level risk factors across different US regions.

## METHODS

### Data Collection

Annual LC mortality rates at the US county level (2015 to 2019) were obtained from the National Cancer Institute's (NCI) SEER Program via SEER*Stat software.[25] County-level risk factors data were collected from several sources, including the American Community Survey (2013–2017), the National Cancer Institute, and the CDC. A detailed description of variables, including units, temporal coverage, and sources, is provided in Table 1.

Predictors were selected on the basis of epidemiological evidence and county-level data availability. Smoking prevalence, PM2.5 levels, and the Environmental Health Index (EHI)—a composite measure of air, water, and land quality exposures—were prioritized as established LC mortality risk factors. Socioeconomic variables (eg, poverty rate and educational attainment) addressed disparities in health care access and health literacy, while demographic (eg, age and race/ethnicity) and geographic factors (eg, rurality and walkability) captured population heterogeneity. Population density contextualized community-level exposure risks.

To ensure robustness, only standardized, publicly available county-level data sets were included. Temporal alignment was prioritized (eg, 2013-2017 American Community Survey estimates for 2015-2019 mortality outcomes), and spatially granular variables (eg, EPA's census tract–level EHI) were aggregated to counties using geographic weighting.

### Statistical Analyses

#### Preprocessing

The primary goal of the preprocessing stage was to address missing values in both dependent and independent variables within the data set. Counties with more than five missing values out of 15 variables were excluded from the analysis. For the remaining entries, missing values were filled by calculating the mean from data of the 20 nearest counties on the basis of the K-nearest neighbors imputation method.[26] Additionally, for variables such as population density, home value, or EHI, which are not measured in percentages, we conducted outlier removal and min-max rescaling processes to standardize the data. These procedures were implemented to improve the consistency and reliability of our analysis.

#### Spatial Analysis

The LC mortality rates were integrated into the shapefile of ArcGIS containing 3,143 counties across the 50 states and the

**TABLE 1.** A List of Dependent and Independent Variables Used in This Study

| Variable Name | Source | Unit | Description |
|---|---|---|---|
| LC mortality rates | NCI | Percent | Annual LC mortality per 100,000 persons (2015-2019) |
| Rural population | ACS | Percent | Percent of population living in rural areas (2013-2017) |
| Age 65 years and older | ACS | Percent | Estimated percent of the population age 65 years and older (2013-2017) |
| Ethnicity—Black | ACS | Percent | Percent of the population that is Black or African American, by single classification of census race (2013-2017) |
| Ethnicity—Hispanic | ACS | Percent | Percent of the population that is Hispanic or Latino (2013-2017) |
| Higher education | ACS | Percent | Estimated percent of the population age 25 years and older with a bachelor's degree, graduate, or professional degree (2013-2017) |
| Poverty rate | ACS | Percent | Estimated percent of all people who are living in poverty (2013-2017) |
| Home value | ACS | Dollar | Estimated median value of an owner-occupied housing unit (2013-2017) |
| Primary care physicians | HRSA | Per 1,000 | The rate of primary care physicians per 1,000 people (2016) |
| PM2.5 air pollution | NEPHTN | μg/cubic meters | Average daily density of fine particulate matter in micrograms per cubic meter (PM2.5; 2018) |
| EHI | EPA | Percent | Average potential exposure to harmful toxins as of 2015, aggregated from the census tract level to the county level |
| Population density | ACS | Per square mile | Number of people per square mile (2013-2017) |
| Walkability index | EPA | Index | Average national walkability index as of 2021, aggregated from the block group level to the county level |
| Smoker rate | CDC places | Percent | Crude percent of current smoking among adults age 18 years and older (2019) |

Abbreviations: ACS, American Community Survey; EHI, Environmental Health Index; EPA, Environmental Protection Agency; HRSA, Health and Resources Service Administration; LC, lung cancer; NCI, National Cancer Institute; NEPHTN, National Environmental Public Health Tracking Network; PCP, Primary Care Physicians.

District of Columbia. First, we visualized the LC mortality rates in a choropleth map and using the Getis-Ord Gi statistic,[27] we pinpointed hotspots denoting regions with elevated rates and coldspots representing areas with lower rates of LC mortality. The Gi* statistic calculates z-scores for each location on the basis of the values of neighboring regions within a specified spatial context, allowing us to detect hotspots—areas with elevated LC mortality rates—and coldspots—regions with lower-than-expected rates. This spatial analysis enabled us to discern localized patterns and trends, offering valuable insights into the geographical variations of LC mortality rates.

*Model Development*

In constructing the predictive models, the response variable was the annual LC mortality rate from 2015 to 2019. The data were randomly divided into training (75%) and testing (25%) sets to develop the predictive models and validate their performance. We used two prominent EL algorithms: RF and gradient boosting regression (GBR), alongside a more traditional LR approach.

RF constructs an ensemble of decision trees, where each tree is trained on a random subset of the data and features. Through bagging and feature randomness, RF mitigates overfitting and enhances model generalization. This EL method excels in capturing complex nonlinear relationships present in our data set, making it well suited for tasks with high-dimensional input features and intricate interactions among predictors.[28,29] GBR, however, operates sequentially by iteratively fitting a series of weak learners, typically decision trees, to the residuals of the preceding model.[30]

Hyperparameters (eg, tree depth, learning rate for GBR, and number of estimators) were optimized using a 5-fold cross-validated grid search on the training set, minimizing root mean squared error (RMSE). This approach ensures that parameter selection avoids overfitting the training data. The final model performance was evaluated on the entirely unseen test set. After model training, predictions of the LC mortality rates were made on the testing set, and the model's accuracy was evaluated using metrics such as coefficient of determination ($R^2$), RMSE, and mean absolute error (MAE). All models were implemented via the Scikit-learn package in Python.

*Improving Explainability*

To identify the important variables contributing to predicting LC mortality rates, Shapley Additive Explanations (SHAP) values were calculated for each variable. SHAP values are a powerful tool for interpreting the output of ML models. They provide a way to understand the contribution of each feature to the model's prediction for a specific instance.[31] We further visualized SHAP summary plots for the six variables with the highest mean SHAP values. This visualization helps interpret the RF model and understand which factors are most important for predicting LC mortality rates in different counties. The summary plot uses individual SHAP value data. The SHAP values can be obtained from either an RF model or a precomputed matrix.

## RESULTS

Of 3,143 counties, 2,820 had nonmissing values for LC mortality with a mean value of 65.5 (SD = 17.9) deaths per 100,000 population.

Figure 1A illustrates the distribution of annual LC mortality rates across the 3,143 US counties for the years 2015 and 2019. Regions in the eastern United States exhibited notably higher rates of LC mortality. Figure 1B presents the outcomes of Getis-Ord Gi* statistics for the clustering of LC mortality rates. The red areas (hotspots) on these maps represent spatial clusters characterized by high LC mortality. As shown, there is a large hotspot area of LC mortality in counties located in the mid-eastern region, while western counties are identified as coldspots.

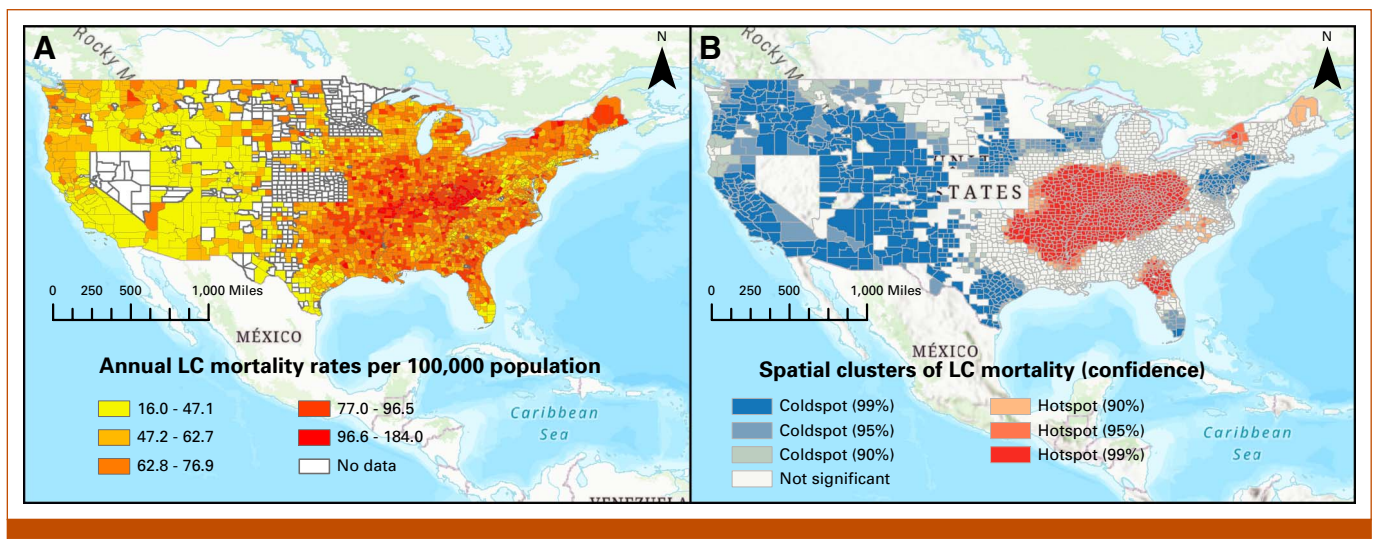

**FIG 1.** Disparities in annual LC mortality rates in US counties using (A) thematic mapping and (B) spatial clustering Getis-Ord Gi* statistic. LC, lung cancer.

ML analysis indicates that, overall, the RF model outperformed the GBR model, exhibiting a higher $R^2$ value of 41.9%, a lower RMSE of 12.8, and a lower MAE of 9.9, compared with the GBR model's $R^2$ value of 38.5%, RMSE of 13.2, and MAE of 10.2. Furthermore, LR demonstrated the poorest performance among the evaluated models, with an $R^2$ value of 31.2%, RMSE of 14.0, and MAE of 10.8, reinforcing the superior performance of ensemble models in this analysis (Fig 2A).

Figure 2B illustrates the significance of each factor in predicting LC mortality rates at the county level in the United States, as determined by SHAP values. The mean SHAP values explain the contribution of each factor to the RF model's predictions. The prevalence of smokers emerges as the most important factor, followed by median home value and the percentage of the Hispanic population. The ranking of these important variables remains consistent across the LR and GBR models.

Figure 3 shows a visualization of SHAP summary plots for the RF model predicting county-level LC mortality rates in the United States. The color on the y-axis represents the SHAP value, which indicates how much a feature contributes to shifting the model's prediction in a certain direction. Positive SHAP values indicate a positive contribution (increasing the predicted LC mortality rates), while negative SHAP values indicate a negative contribution (decreasing the predicted LC mortality rates). For example, in the smokers rate plot, we see that counties with higher smoking rates (on the right side of the plot) tend to have higher predicted LC mortality rates (indicated by the positive SHAP values). Conversely, counties with lower smoking rates (on the left side of the plot) tend to have lower predicted LC mortality rates (indicated by the negative SHAP values). The vertical spread of the points represents the feature's impact on different data instances. In other words, it shows how the effect of a feature can vary depending on the other features within a given county. In summary, the analysis of the top six variables through SHAP summary plots reveals that the rate of smoking, air pollution, and population density have a direct influence on LC mortality rates. Conversely, factors such as median home value, Hispanic ethnicity, and levels of higher education appear to mitigate LC mortality rates.

Figure 4 compares the spatial distribution of the top six predictors with LC mortality disparities. Smoking prevalence hotspots significantly overlapped with LC mortality hotspots, particularly in the mid-eastern United States. By contrast, counties with higher percentages of Hispanic populations and higher median home values aligned with LC coldspots, suggesting potential protective effects. Coldspots of higher education were also more prevalent in areas with higher LC mortality. Other variables, such as population density and air pollution, did not exhibit clear spatial associations with LC mortality patterns.

## DISCUSSION

Our investigation focused on comparing the predictive performance of explainable EL models with their LR counterparts for county-level LC mortality rates in the United States. The results strongly supported the ensemble models. Furthermore, the findings of this study shed light on the complex interplay between various sociodemographic factors and LC mortality rates at the county level in the United States. As a secondary objective, we explored the geographic disparities in LC mortality rates, revealing distinct spatial patterns, with higher rates predominantly concentrated in the eastern regions. These observed disparities highlight the critical role of local contextual factors in developing effective public health interventions.

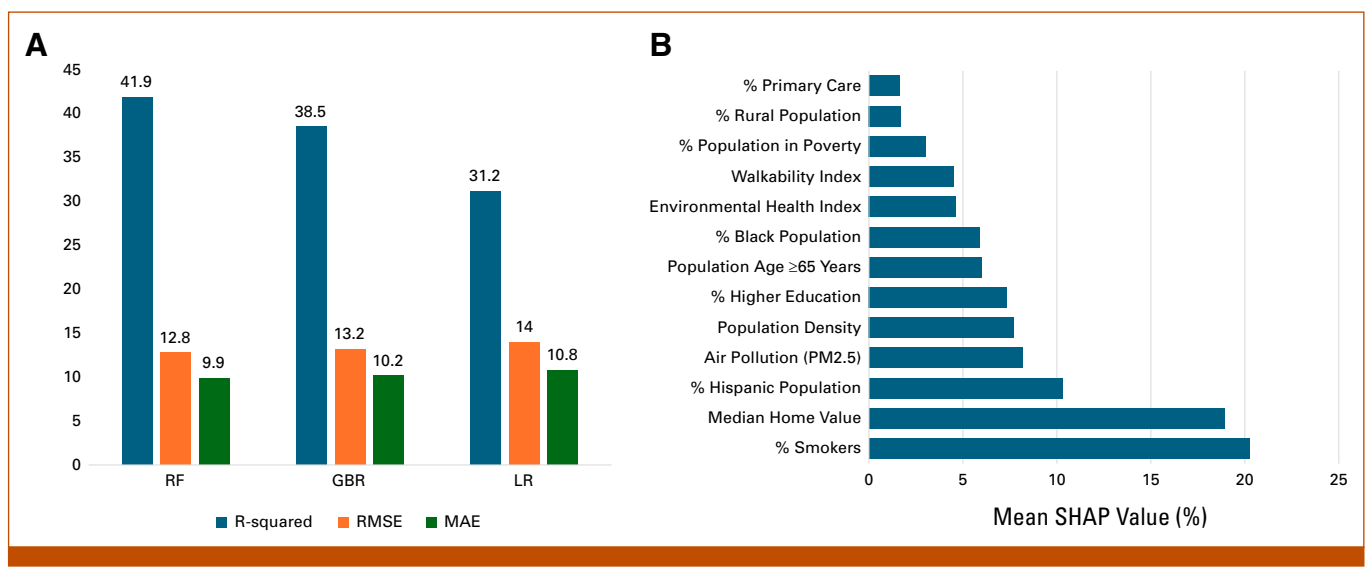

**FIG 2.** (A) Performance of RF, GBR, and LR models in prediction of county-level LC mortality rates, and (B) SHAP values explaining factors influencing LC mortality rates predicted by a RF model across US counties. GBR, gradient boosting regression; LC, lung cancer; LR, linear regression; MAE, mean absolute error; RF, random forest; RMSE, root mean squared error; SHAP, Shapley Additive Explanations.

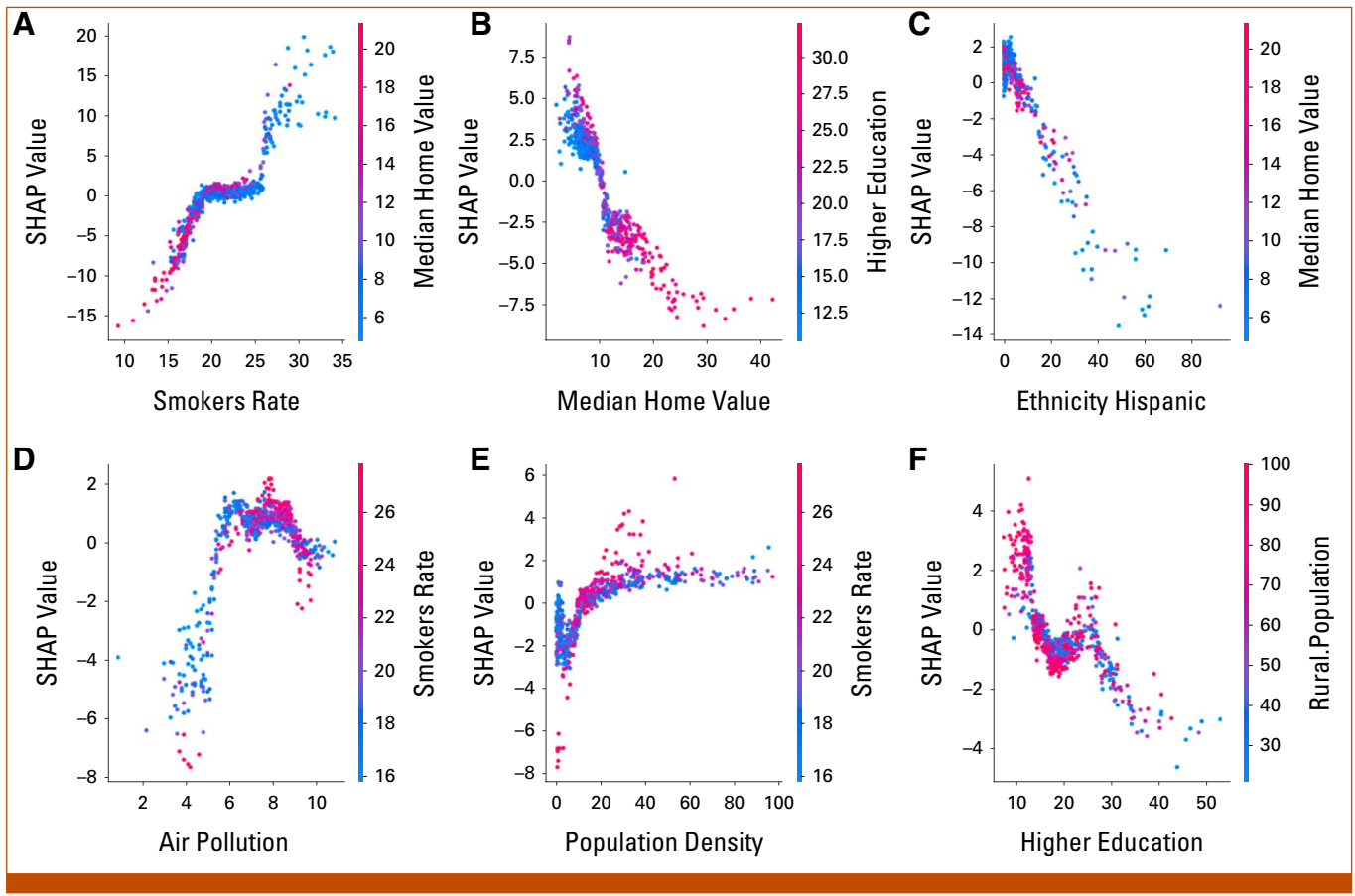

**FIG 3.** SHAP summary plots for the top six features influencing LC mortality predictions by a RF model: (A) smokers rate, (B) median home value, (C) ethnicity Hispanic, (D) air pollution, (E) population density, and (F) higher education; the color indicates a feature's impact (positive for higher predicted mortality, negative for lower). Wider spread reflects variation in a feature's effect depending on other factors. LC, lung cancer; RF, random forest; SHAP, Shapley Additive Explanations.

The superiority of the RF model in predicting LC mortality rates highlights the effectiveness of ensemble ML techniques in capturing the nonlinear relationships inherent in the data set. This finding aligns with previous studies that have demonstrated the effectiveness of ensemble methods in handling complexities within health outcome data, particularly disparities in health outcomes.[19,32-34] The SHAP summary plots also show these nonlinear relationships between factors and outcomes, explaining why LR underperformed compared with ML models such as RF and GBR, which can account for nonlinear relationships. Furthermore, ensemble methods such as RF aggregate predictions from multiple decision trees, effectively averaging out individual tree errors. Consequently, this reduces the model's variance, leading to more robust and generalizable predictions compared with single models such as LR. The RF model's superiority in this study provides valuable feature importance scores, indicating which factors most significantly influence LC mortality rates. This deeper understanding of the driving factors enables targeted interventions.

In accordance with well-established risk factors for LC, our analysis using the RF model identified smoking prevalence as the factor exerting the strongest influence on county-level LC mortality rates. SHAP summary plots demonstrated a consistent positive association, with counties exhibiting higher smoking rates showing disproportionately elevated predicted LC mortality. Although individual-level risk factors may not always translate directly to ecological analyses, our findings align with population-level studies affirming smoking as a key driver of LC outcomes.[5,19,34] This relationship persists even when accounting for contextual variables such as environmental exposures and socioeconomic disparities, underscoring smoking's unique explanatory power in geospatial models of LC mortality. Importantly, the broader public health implications of smoking extend beyond LC, as evidenced by its linkage to other adverse health trends across US counties.[32,33,35] These results emphasize the urgent need for targeted smoking cessation policies in high-risk regions, leveraging ML's capacity to prioritize actionable variables in complex public health data sets.

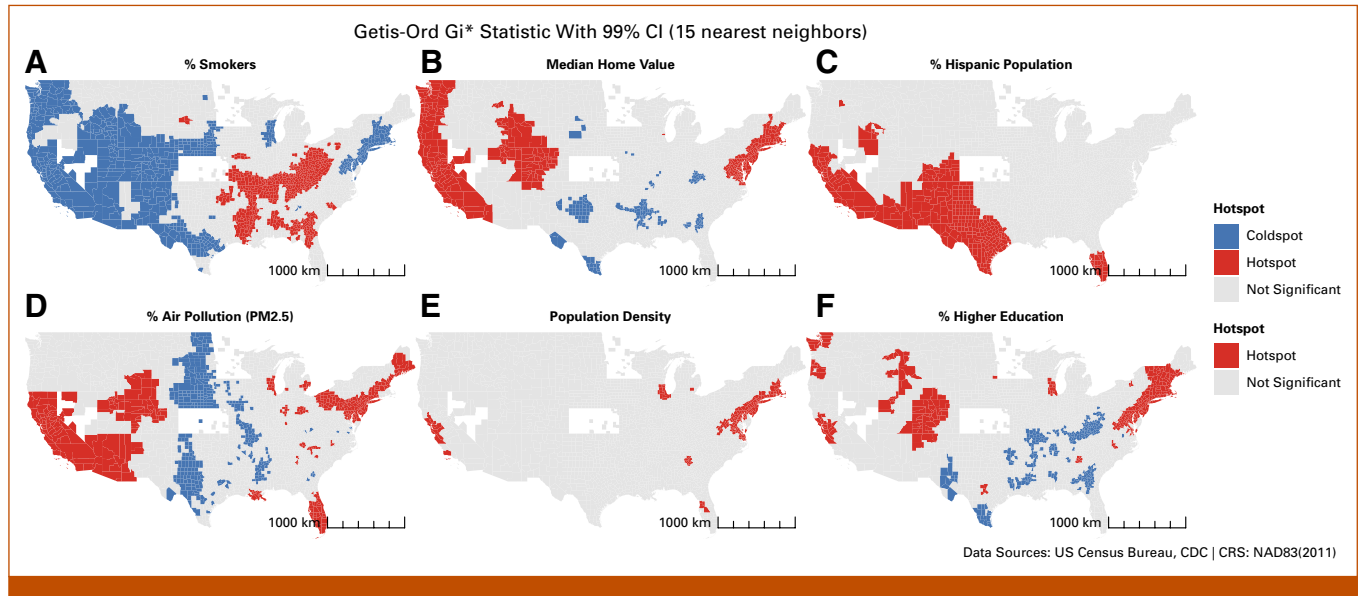

**FIG 4.** Spatial clusters of the top six county-level features influencing LC mortality predictions using Getis-ord Gi*: (A) % smokers, (B) median home value, (C) % Hispanic population, (D) % air pollution (PM2.5), (E) population density, and (F) % higher education. LC, lung cancer.

Shreves et al[36] investigated the geographic patterns of LC mortality and cigarette smoking in the United States. Their study, consistent with our findings, revealed that spatial patterns of ever-smoking were generally associated with LC mortality rates. Specifically, these rates were elevated in the Appalachian region and lower in the West for both sexes, similar to our spatial clustering. This underscores the importance of public health initiatives aimed at smoking cessation to reduce LC risk. However, Shreves et al also observed unexplained geographic variations in mortality rates, emphasizing the need to consider factors beyond smoking.[36] Their analyses identified certain US counties where factors other than smoking may be driving LC mortality. Notably, our RF model emphasized the significance of median home value, Hispanic ethnicity, air pollution, and higher education as important factors associated with LC mortality. These findings highlight the critical role of public health interventions targeting smoking cessation, air quality improvement, and addressing socioeconomic disparities in LC risk reduction strategies. Additionally, our resulting spatial pattern in LC mortality resembles the patterns of other cancer mortalities. Dong et al conducted a study of cancer mortality disparities among US counties from 2008 to 2019.[37] Their findings support that mid-eastern counties exhibit elevated rates of cancer mortality, which aligns with our research. They identified smoking, receipt of Supplemental Nutrition Assistance Program (SNAP) benefits, and obesity as the primary factors contributing to cancer mortality, underscoring the pervasive impact of smoking on not only LC but also across all cancer types.

Although this study advances our understanding of county-level LC mortality determinants, several limitations warrant consideration. First, although our model incorporated robust predictors such as smoking prevalence, air pollution, and socioeconomic factors, the ecological study design inherently limits causal inference and introduces the risk of ecological fallacy. County-level aggregation may mask individual-level variations in risk factors (eg, undiagnosed comorbidities, genetic predispositions, or occupational exposures), which could not be assessed because of data availability constraints. Second, key variables linked to LC mortality in previous research—including occupational hazards (eg, asbestos and radon), dietary patterns, and indoor air pollution—were excluded because of a lack of standardized, county-level data. For instance, workplace exposures and household fuel combustion practices, which disproportionately affect rural and industrial communities, remain underrepresented in public data sets. Similarly, genetic susceptibility data, often inaccessible at the population level, were not included. Third, reliance on aggregated data introduces potential temporal misalignment. For example, mortality rates (2015-2019) were paired with predictor variables spanning 2013-2021, which may not fully account for latency periods between exposures and LC outcomes. Although this temporal gap could introduce some uncertainty, previous studies suggest that environmental and socioeconomic factors, including air pollution levels and health care access metrics, generally remain stable over short periods.[3,4] Finally, unmeasured confounders, such as regional differences in screening adherence or health care quality, could influence mortality rates but were not captured in our analysis.

Despite these limitations, our findings provide a foundational understanding of LC mortality patterns using transparent, publicly available data. Future studies could address these gaps by integrating individual-level electronic health records, longitudinal environmental monitoring, or geospatial proxies for localized exposures (eg, industrial zoning maps).

In conclusion, this study investigated LC mortality disparities in the United States, using spatial, statistical, and ML methodologies. The explainable ML models constructed by the RF algorithm demonstrated superiority over GBR and LR models. Counties in the mid-eastern region exhibited significantly higher rates of LC mortality, underscoring the need for policy action. Furthermore, our SHAP analysis of the RF model identified smoking as the most influential factor in LC mortality rates, highlighting an opportunity for educational and legislative interventions in high-risk regions. The spatial patterns and influential factors identified offer valuable insights for policymakers, health care practitioners, and researchers seeking to devise targeted interventions to reduce LC mortality. This spatial heterogeneity requires geographically tailored interventions that address the specific socioeconomic and environmental determinants of LC in each region. Sustained research and intervention initiatives are crucial for effectively addressing LC, particularly in regions with high mortality rates.

### AFFILIATIONS

[1]Department of Pediatrics, Center for Biomedical Informatics, College of Medicine, University of Tennessee Health Science Center, Memphis, TN
[2]Pharmaceutical Research Center, Pharmaceutical Technology Institute, Mashhad University of Medical Sciences, Mashhad, Iran
[3]Department of Radiation Oncology, College of Medicine, University of Tennessee Health Science Center, Memphis, TN

### CORRESPONDING AUTHOR

Arash Shaban-Nejad, PhD, MPH; e-mail: ashabann@uthsc.edu.

### AUTHOR CONTRIBUTIONS

**Conception and design:** Soheil Hashtarkhani, Benyamin Hoseini, David L. Schwartz, Arash Shaban-Nejad
**Financial support:** Arash Shaban-Nejad
**Administrative support:** Arash Shaban-Nejad
**Provision of study materials or patients:** Arash Shaban-Nejad
**Data analysis and interpretation:** All authors
**Manuscript writing:** All authors
**Final approval of manuscript:** All authors
**Accountable for all aspects of the work:** All authors

### AUTHORS' DISCLOSURES OF POTENTIAL CONFLICTS OF INTEREST

The following represents disclosure information provided by authors of this manuscript. All relationships are considered compensated unless otherwise noted. Relationships are self-held unless noted. I = Immediate Family Member, Inst = My Institution. Relationships may not relate to the subject matter of this manuscript. For more information about ASCO's conflict of interest policy, please refer to www.asco.org/rwc or ascopubs.org/cci/author-center.

Open Payments is a public database containing information reported by companies about payments made to US-licensed physicians (Open Payments).

**David L. Schwartz**
**Research Funding:** National Cancer Institute, Elekta (Inst)
**Travel, Accommodations, Expenses:** National Comprehensive Cancer Network
**Other Relationship:** NRG Cooperative Research Group

**Arash Shaban-Nejad**
**Consulting or Advisory Role:** Haleon
**Travel, Accommodations, Expenses:** Haleon

No other potential conflicts of interest were reported.